\documentclass[10pt,twocolumn,letterpaper]{article}
\usepackage{graphicx}
\usepackage{subcaption} 

\usepackage{cvpr}              
\usepackage{makecell}
\usepackage{amsthm}
\usepackage{pifont}
\usepackage[accsupp]{axessibility}

\usepackage{enumitem}  
\usepackage{adjustbox}
\usepackage{booktabs}  
\definecolor{greyL}{RGB}{235,235,235}
\usepackage{tcolorbox}
\usepackage{multirow}
\tcbset{
  highlightbox/.style={
    colback=greyL, 
    colframe=black, 
    boxrule=1pt, 
    arc=5pt, 
    boxsep=5pt, 
    left=5pt, right=5pt, top=0pt, bottom=0pt, 
    width=0.47\textwidth, 
    fontupper=\itshape 
  }
}

\newcommand{\method}{FedRG}

\definecolor{cvprblue}{rgb}{0.21,0.49,0.74}
\usepackage[pagebackref,breaklinks,colorlinks,allcolors=cvprblue]{hyperref}

\def\paperID{42633} 
\def\confName{CVPR}
\def\confYear{2026}

\title{FedRG: Unleashing the Representation Geometry for Federated Learning with Noisy Clients}

\author{
Tian Wen$^{1}$\thanks{Equal contribution.} \quad
Zhiqin Yang$^{2}$\footnotemark[1] \quad
Yonggang Zhang$^{4}$ \quad
Xuefeng Jiang$^{1}$ \quad
Hao Peng$^{3}$ \\
Yuwei Wang$^{1} $\thanks{Corresponding author: ywwang@ict.ac.cn,bhanml@comp.hkbu.edu.hk} \quad
Bo Han$^{2}$\footnotemark[2] \\
$^{1}$Institute of Computing Technology, Chinese Academy of Sciences \\
$^{2}$TMLR Group, Hong Kong Baptist University \quad
$^{3}$Beihang University \\
$^{4}$The Hong Kong University of Science and Technology  \\
}

\begin{document}
\maketitle
\begin{abstract}
Federated learning (FL) suffers from performance degradation due to the inevitable presence of noisy annotations in distributed scenarios. 
Existing approaches have advanced in distinguishing noisy samples from the dataset for label correction by leveraging loss values.
However, noisy samples recognition relying on scalar loss lacks reliability for FL under heterogeneous scenarios. 
In this paper, we rethink this paradigm from a representation perspective and propose \method~(\textbf{Fed}erated under \textbf{R}epresentation \textbf{G}eometry), which follows \textbf{the principle of ``representation geometry priority''} to recognize noisy labels. 
Firstly, \method~creates label-agnostic spherical representations by using self-supervision. It then iteratively fits a spherical von Mises-Fisher (vMF) mixture model to this geometry to capture semantic clusters. This geometric evidence is integrated with a semantic-label soft mapping mechanism to derive a distribution divergence between the label-free and annotated label-conditioned feature space, which robustly identifies noisy samples and updates the semantic-label mappings with the newly separated clean dataset. Lastly, we employ an additional personalized noise absorption matrix on noisy labels to achieve robust optimization.
Extensive experimental results demonstrate that \method~ outperforms state-of-the-art methods for FL  with data heterogeneity under diverse noisy clients scenarios. Our code is publicly available at \url{https://github.com/Tianjoker/FedRG}.
\end{abstract}    
\section{Introduction}
\label{sec:intro}
\begin{figure}[tbp]
    \centering
\includegraphics[width=1\linewidth]{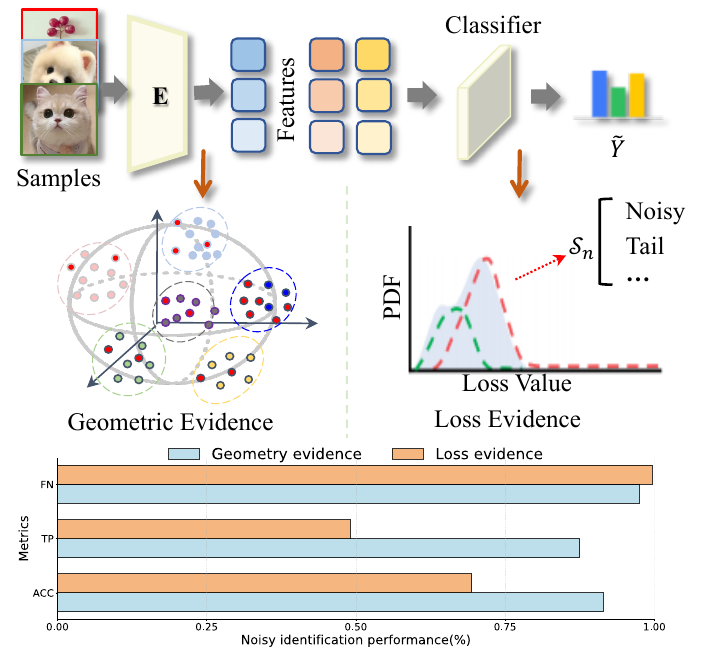}
    \caption{Comparison of noise identification performance between the existing method and \method~under severe data heterogeneity. The existing methods mainly identify based on the distribution of scalar loss values. In heterogeneous scenarios, the loss value alone becomes unreliable, while \method~provides more robust spatial geometry to filter out noise samples. Higher FN (False Negatives) but lower TP (True Positives) indicate that noise identification based on loss value wrongly classifies many correct labels as noise labels. The red circle denotes that the noisy samples we want to identify based on the geometric evidence.}\label{fig:motivation_fig}
    \vspace{-0.6cm}
\end{figure}

Federated learning (FL)~\cite{kairouz2021advances,yang2019federated} enables collaborative model training across multiple data sources without centralizing raw data. In FL, local models are trained on client data and subsequently aggregated on the server to obtain a robust global model~\cite{fedavg,li2021survey}. Nevertheless, the distributed architecture introduces critical challenges. Specifically, the data distribution across clients is inherently heterogeneous, also known as the Non-IID problem~\cite{mendieta2022local,wang2024aggregation,yangfedgps}. Furthermore, the variability or scarcity of labeling experts often results in the presence of noisy labels within the heterogeneous datasets~\cite{jiangfedclean,peijian}. These combined factors critically hinder the performance and practical deployment of FL systems~\cite{Fednoisy,yang2023fedfed}.

Existing methods have made notable progress in enhancing the robustness of FL against noisy labels~\cite{DBLP:journals/tvt/YangQWZZ22,fedcorr,fedfixer}.
Some approaches adopt robust aggregation or training strategies to alleviate the adverse impact of noisy samples~\cite{FedNoRo, symmetricCE, lycklama2023rofl,zeng2024fedes}. In traditional machine learning, loss values serve as the reliable indicator for identifying noisy samples~\cite{co-teaching,Gui2021TowardsUD,xiasample}. 
Drawing from this intuition, several FL studies distinguish between noisy and clean samples by leveraging insights from conventional label noise research~\cite{FedNoRo,fedcorr}. 
For example, some methods improve FL robustness in noisy environments by exploiting the distribution of losses~\cite{FedNoRo,zhou2024federated}. Other works utilize loss values in a more granular manner. For instance, FedCorr~\cite{fedcorr} employs a Gaussian mixture model (GMM) to separate noisy and clean data based on loss values, while FedClean~\cite{jiangfedclean} assesses the consistency between annotated and inferred labels using the same metric.

Built upon existing methods, it is natural that we raise a key \textit{\textbf{Question 1:} Can the loss value be reliably used to identify noisy labels in heterogeneous FL scenarios?}
Recall that label skew represents a primary form of heterogeneity in FL. This local label imbalance often results in a long-tailed distribution within client datasets~\cite{zhang2022federated,diao2023towards}. Training on such imbalanced data introduces a strong bias in the local model toward dominant classes, impairing the classifier's ability to accurately discern samples from rare classes~\cite{zhang2023noisy}. As a result, tail-class samples inherently exhibit higher loss values, irrespective of label accuracy~\cite{baik2024distribution,jiang2022delving}. Apart from the loss value, we turn our attention to the representation space. 
\textbf{To answer \textit{Question 1},} we shift our focus from mere loss values to the representation space itself. Furthermore, we propose the representation geometry priority principle to identify noisy labels: the intrinsic representation geometry, which is independent of labels, should align consistently with the geometry shaped by the constraints of annotated labels. We illustrate the differences between these two strategies in Fig.~\ref{fig:motivation_fig}.

As a consequence, emerging another \textit{\textbf{Question 2}: How to measure the geometric representation divergence with and without the annotated labels?}
Revisit the self-supervised methods that have attracted substantial attention, owing to their capacity to extract data representations without dependence on labels~\cite{simclr,liu2021self, psaltis2023fedlid}. \textbf{To address \textit{Question 2},} we leverage instance discrimination to ensure that the features of local data are distributed across a hypersphere~\cite{wang2020understanding}. To model the intrinsic semantic structure on this hypersphere, we adopt the von Mises-Fisher (vMF) distribution~\cite{gopal2014mises,hasnat2017mises}. Then the measurement of geometric divergence is formulated through a cross-validation between two independently derived cluster distributions within vMF: label-free
geometry evidence $\Gamma$ and annotated label-dependent geometry evidence $\text{B}$.
Drawing the intuition from above analysis, we propose \textbf{\method}~(\textbf{Fed}erated under \textbf{R}epresentation \textbf{G}eometry), to tackle noisy clients under data heterogeneity in FL. Based on the recognized noisy labels, we train the whole model coupled with the noise absorption matrix which estimates a personalized noise transition probability for each client.
Overall, our contributions are summarized as follows:

\begin{itemize}
    \item We rethink the paradigm of existing federated learning with noisy clients beyond the loss value. From the representation perspective, we propose the principle of representation geometry priority to guide the noise identification procedure.
    \item We propose \method~ to distinguish noisy samples based on the geometric evidence under the vMF distribution. We draw the inspiration that a clean sample's intrinsic feature clustering should align with the clustering pattern implied by its annotated label, while mislabeled samples exhibit geometric inconsistency.
    \item We conduct comprehensive experiments across multiple datasets with four noisy scenarios, including two that arise specifically from the distributed setting. The experimental results consistently demonstrate the efficacy of \method. Further ablation studies also show the effectiveness of each component.
\end{itemize}

\section{Related Work}
\label{sec:formatting}

\subsection{Federated Learning}
In recent years, Federated Learning (FL) \cite{fedSummary,tanhao} has been widely studied and applied in privacy-sensitive domains such as medical analysis \cite{fedeye,FedNoRo} and remote sensing \cite{fedrs}. 
FedAvg \cite{fedavg} aggregates local models through weighted averaging and converges efficiently under the IID assumption.
To cope with the Non-IID data (i.e. statistical heterogeneity), FedProx ~\cite{fedprox} introduces a proximal term while SCAFFOLD~\cite{scaffold} uses control variates for variance reduction. MOON~\cite{moon} conducts representation alignment via contrastive learning \cite{simclr}. 
While these methods demonstrate effectiveness on clean datasets, their robustness diminishes under noisy labels, which is a common concern for real-world scenarios. 

\subsection{Noisy Label Learning} 
Noisy label learning (NLL) \cite{nll} aims to mitigate the negative impact of noisy labels and train models in centralized scenarios. Regularization methods are commonly utilized by designing robust loss functions \cite{MAE,GCE,symmetricCE}. 
Symmetric cross-entropy~\cite{symmetricCE} augments CE with reverse cross-entropy (RCE). Mean Absolute Error (MAE)~\cite{MAE}  provides formal robustness properties relative to CE. Since these approaches operate entirely at the client side, they can be incorporated into FL by substituting the local training objective, without altering the FL protocol. 

\subsection{Noisy Labels in Federated Learning}
Centralized noise-robust training is difficult to deploy in FL due to privacy and communication constraints \cite{elr}. Prior federated methods facing label noise fall into two streams \cite{Fednoisy}: \emph{Client-level assessment} scores the reliability of each client and adapts training accordingly. FedNoRo~\cite{FedNoRo} separates clean and noisy clients using a GMM-based criterion~\cite{gmm} and then aligns unreliable clients via knowledge distillation~\cite{kd}; 
\emph{Sample-level assessment} instead targets example credibility on-device. 
Despite their efficacy, these approaches share notable weaknesses. Most client or data detectors hinge on the \emph{small-loss} heuristic \cite{fedcorr,FedNoRo}, which is unreliable. Deep models eventually memorize mislabeled data, driving their losses down and causing noisy samples to be kept, while truly hard but correctly labeled examples often retain large losses and are discarded~\cite{fedpca}. Additional related work is provided in Appendix Sec.6.

Motivated by these observations, we propose a federated noise-robust framework that, from a geometric representation perspective, identifies clean and noisy samples to guide robust optimization. Unlike previous approaches that rely solely on small-loss heuristics~\cite{FedNoRo,fedcorr}, our method exploits geometric relationships within the feature space to capture structural consistency, enabling effective learning under heterogeneous federated conditions.

\begin{figure*}[t]
\centering
\includegraphics[width=1\linewidth]{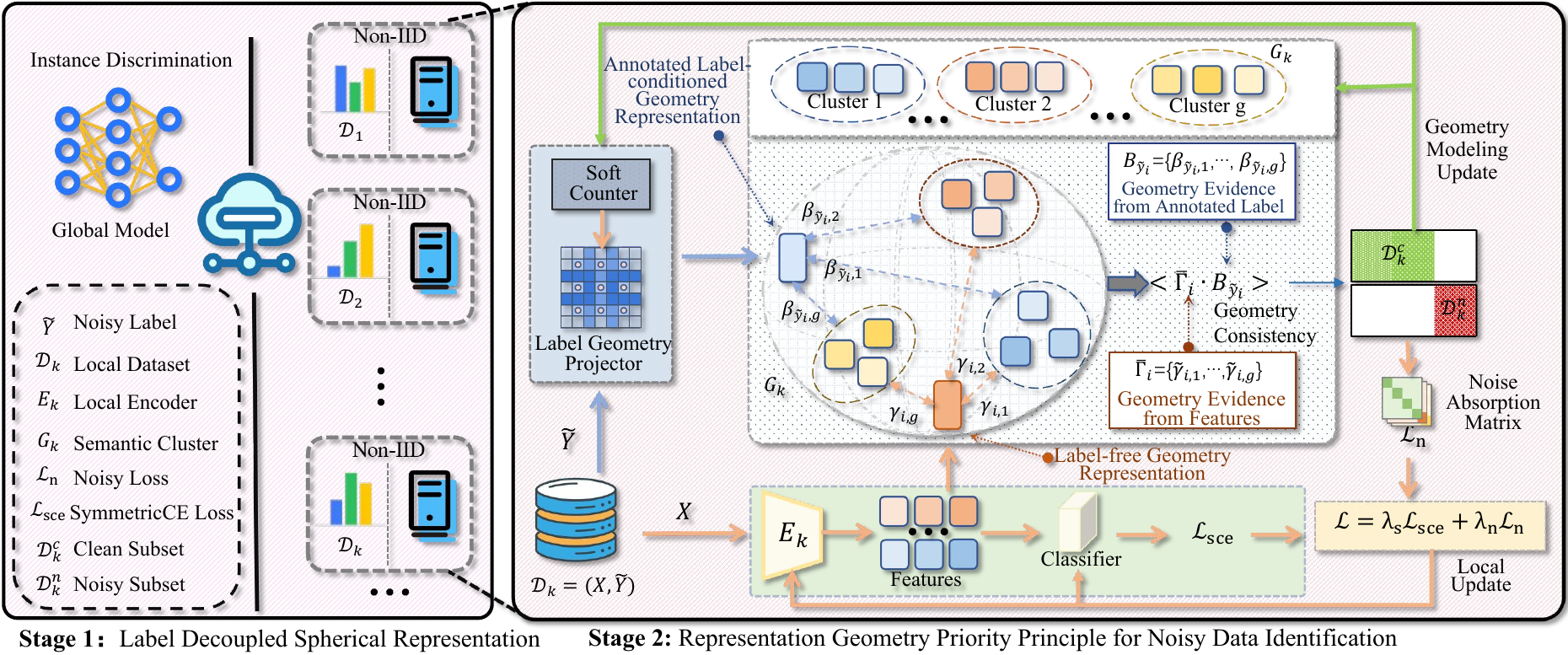}
\caption{Overview of \method. The label decoupled spherical representation stage gets the hypersphere of samples. Then, we follow the representation geometry priority principle to split the clean and noisy samples based on the distribution divergence in the vMF distribution. The pseudocode overview of \method~is provided in Appendix Sec.2}
\label{fig:overview}
\vspace{-0.6cm}
\end{figure*}

\section{Methodology}
\subsection{Preliminaries}
\paragraph{Federated Learning.}
We investigate a standard federated learning framework in which $K$ distributed devices (clients) collaboratively train a $C$-class classification model coordinated by the central server. Each client $k\in[K]$ holds a private dataset $\mathcal{D}_k \;=\; \{(x_i^{(k)},\, y_i^{(k)})\}_{i=1}^{n_k},x\in\mathcal{X},\ y\in\mathcal{Y}=\{1,\dots,C\}$. Without loss of generality, we explicitly allow statistical heterogeneity across clients, where local samples are drawn i.i.d. from client-specific distribution $P_k$ on $\mathcal{X}\times\mathcal{Y}$ and these distributions differ across clients (i.e., Non-IID setting). The server aims to learn a global model parameterized as $\theta\in\Theta$ with a predictor $f_\theta:\mathcal{X}\to\Delta^{C-1}$, by minimizing the following objective:
\begin{equation}
\label{eq:fed-objective}
\min_{\theta\in\Theta}\ \ \mathcal{L}(\theta)
\;=\; \sum_{k=1}^K\omega_t^{(k)}\ \mathbb{E}_{(x,y)\sim P_k}\!\Big[\ell\big(f_\theta(x),\,y\big)\Big],
\end{equation}
where $\ell$ is the loss function and $\omega_t^{(k)}$ determines the aggregated weight of the $k$-th client. During each communication round $t\in \{1,\ldots,T \}$, the server broadcasts the latest global parameters $\theta_t$ to a subset of clients $\mathcal{S}_t\subseteq[K]$. Each participating client performs $E$ local epochs of stochastic optimization on private data to obtain the updated model $\theta_{t+1}^{(k)}$, and then the server aggregates $\theta_{t+1}
\;=\;
\sum_{k\in\mathcal{S}_t}\omega_t^{(k)}\,\theta_{t+1}^{(k)}$ for the next training round.

\paragraph{Federated Label Noise.}
In practical FL deployments, local annotations are often imperfect due to weak supervision or crowdsourcing. 
Following most previous works \cite{FedNoRo, fedrom}, we adopt the instance-independent noise model and describe label corruption by a transition kernel $\mathcal{C}(\tilde{y}\mid y)$, where $\sum_{\tilde{y}=1}^{C}\mathcal{C}(\tilde{y}\mid y)=1$ and $\mathcal{C}(\tilde{y}\mid y)\in[0,1]$. For client $k$, the observed label is drawn as $\tilde{y}\sim\mathcal{C}(\cdot\mid y)$, and the local training set is $\tilde{\mathcal{D}}_k=\{(x_i^{(k)},\tilde{y}_i^{(k)})\}_{i=1}^{n_k}$. 

We study two noise types in federated systems. (i) Globalized noise: a \emph{single} kernel $\mathcal{C}(\tilde{y}\mid y)$ governs flips for all clients; equivalently, $p(\tilde{y}=j\mid y=i)=\mathcal{C}(j\mid i)$ is identical throughout the federation, where $p(\tilde{y}=j\mid y=i)$ is the probability of flipping from the true label $y$ of class $i$ to the wrong label  $\tilde{y}$ of class $j$. (ii) Localized noise: each client $k$ has the personalized kernel $\mathcal{C}_k(\tilde{y}\mid y)$ that operates only within its local label support $\mathcal{Y}_k\subseteq\mathcal{Y}$. In particular, $\mathcal{C}_k(\tilde{y}\mid y)=0$ whenever $\tilde{y}\notin\mathcal{Y}_k$. This captures realistic client-specific biases and distinguishes our setting from ordinary label-noise assumptions that allow flips to any global class regardless of local availability. 

For each of the noise types, we consider the symmetric flipping and pairwise flipping (i.e., pairflip) methods, aligning with prior work~\cite{Fednoisy}. In the symmetric flipping, each true label can be randomly flipped to any other admissible alternative class with equal probability. By contrast, pairflip indicates that each class is primarily confused with one specific class. More details are provided in Sec. \ref{implement_details}.

\subsection{Motivation}

Practical federated learning faces severe non-IID data and heterogeneous label noise, under which the prevalent small-loss heuristic \cite{fedcorr, fedrom} becomes unreliable, since losses conflate mislabels, rare yet clean examples, and domain shift. Moreover, noisy warm-up can already distort the representations, weakening subsequent loss-based filtering. We therefore advocate the representation geometry priority principle, where robust assessment of sample cleanliness arises from evidence in the intrinsic representation geometry rather than that from the easily contaminated prediction space. This motivates the two-stage design of \method, illustrated in Fig. \ref{fig:overview}, which first learns label-decoupled spherical representations and then performs geometry-based noise identification and robust optimization.

\subsection{Label-Decoupled Spherical Representation}
\label{warmup}
To instantiate the representation geometry priority principle, we require an initialization that learns stable hyperspherical representations without using annotated labels and whose inductive bias matches the spherical modeling used later. We therefore adopt instance discrimination (i.e., SimCLR) for Stage-I federated pretraining~\cite{wang2020understanding,simclr}. Each sample \(x\) is stochastically augmented twice and encoded into a unit-norm representation on the hypersphere as
$z \;=\; \frac{f_\theta(x)}{\left\| f_\theta(x) \right\|}\; \in\; \mathbb{S}^{d-1},$ which is optimized with the symmetric NT-Xent (InfoNCE) objective over \(2b\) views, where \(b\) is the batch size. For a positive pair \((i,i^+)\), the training loss is
\begin{equation}
\mathcal{L}_{\text{SimCLR}}
=\frac{1}{2b}\sum_{i=1}^{2b}
-\log\frac{\exp\!\left(z_i^\top z_{i^+}/\tau\right)}
{\sum_{\substack{a=1\\ a\neq i}}^{2b}\exp\!\left(z_i^\top z_a/\tau\right)},
\end{equation}
where the temperature \(\tau\) controls the sharpness of contrastive similarities on \(\mathbb{S}^{d-1}\). This objective encourages alignment of positive views with similar semantics while promoting feature uniformity on the sphere, yielding a stable geometric substrate in which fine-grained semantic neighborhoods emerge as directional concentrations and samples with weak directional structure receive less concentrated geometric evidence.

After \(T_1\) rounds of label decoupled spherical representation under federated training, clients can obtain the well-conditioned spherical representation manifold on \(\mathbb{S}^{d-1}\), which is subsequently used for the vMF-based geometric modeling in Sec.~\ref{Representation Geometry Prior}.

\subsection{Representation Geometry Priority Principle}
\label{Representation Geometry Prior}

Given the label-decoupled spherical representations produced in Sec. \ref{warmup}, we now instantiate the representation geometry priority principle by placing an explicit probabilistic model on the hypersphere \(\mathbb{S}^{d-1}\). The key idea is that assessment of sample cleanliness should arise from the intrinsic geometry of these representations.

\subsubsection{Geometric Representation via vMF Distribution}
On the SimCLR induced hyperspherical manifold, we expect semantic regularities to manifest as directional concentrations: samples sharing stable, view-invariant factors (e.g., textures or poses) should aggregate around a few preferred directions, while samples with weak or unstable directional support receive larger background mass. To turn this intuition into a probabilistic model that respects the representation geometry, we therefore seek a distribution on \(\mathbb{S}^{d-1}\) that can represent such directional modes together in a geometrically consistent way. The von Mises–Fisher (vMF) distribution is a natural choice for the directional component.
It is defined on the unit hypersphere and smoothly interpolates between a sharply peaked mode around a mean direction and the uniform distribution when no directional preference exists~\cite{banerjee2005clustering,gopal2014mises}. We therefore model the distribution of normalized representations as a mixture of vMF components and a uniform background. 
Formally, let $z\!\in\!\mathbb S^{d-1}$ denote the normalized representation, we assume:
\begin{equation}
p(z)\;=\;\pi_0\,U(z)\;+\;\sum_{g=1}^{G}\pi_g \,\mathrm{vMF}\!\left(z\,\middle|\,\mu_g,\kappa_g\right),
\label{eq:vmf_mixture}
\end{equation}
where $U(z)$ is the uniform density on $\mathbb{S}^{d-1}$ and $\{\pi_g\}_{g=0}^G$ are mixture weights satisfying $\sum_{g=0}^G \pi_g = 1$. Each vMF component corresponds to a semantic cluster (indexed by $g$) on the sphere, capturing a directionally concentrated mode induced by stable, view-invariant factors, while the uniform component absorbs geometry-level outliers that do not align with any semantic mode. Importantly, clusters are \emph{label-agnostic}, as they capture geometry-driven regularities that may refine a class into sub-modes or, conversely, span cross-class motifs. This decoupling plays a key role under noisy supervision, and in the next step, we turn this mixture into sample-wise geometric evidence.

\subsubsection{Label-free Geometry Evidence.}
Given the vMF mixture in Eq.~(\ref{eq:vmf_mixture}), we first extract label-free geometric evidence in the form of posterior responsibilities over semantic clusters. Let $p_0(z){=}U(z)$ and $p_g(z){=}\mathrm{vMF}(z\mid\mu_g,\kappa_g)$ for $g{\ge}1$, with mixture weights $\{\pi_g\}_{g=0}^G$. For each sample of a client, the posterior responsibility of belonging to semantic cluster $g \in \{0,1,\dots,G\}$ is:
\begin{equation}
\gamma_{i,g}\;=\;\frac{\pi_g\,p_g(z_i)}{\sum_{h=0}^{G}\pi_h\,p_h(z_i)},
\label{eq:resp}
\end{equation}
which constitute our initial label-free geometric evidence. 
 
To further exploit semantic stability, we refine these responsibilities using multi-view consistency. For each training sample $x_i$, we utilize the standard SimCLR-style stochastic augmentations and obtain unit-norm features $z_{i,1}, z_{i,2} \in \mathbb{S}^{d-1}$. Their agreement yields a scalar consistency score, which is mapped to a precision-like tempering factor $r_i$; this factor is then used to flatten responsibilities for low-consistency samples and to preserve sharp assignments for well-aligned ones. The full augmentation pipeline and the resulting tempered likelihoods are detailed in Appendix Sec.3.

Using these tempered likelihoods, the responsibilities become a refined vector $\tilde{\gamma}_{i,g}$ as:
\begin{equation}
\tilde{\gamma}_{i,g}
=
\begin{cases}
\displaystyle
\frac{\pi_0\,\tilde{p}_{0}(z_i;r_i)}{\pi_0\,\tilde{p}_{0}(z_i;r_i)\;+\;\sum_{h=1}^{G}\pi_h\,\tilde{p}_{h}(z_i;r_i)}, & g=0,\\[10pt]
\displaystyle
\frac{\pi_g\,\tilde{p}_{g}(z_i;r_i)}{\pi_0\,\tilde{p}_{0}(z_i;r_i)\;+\;\sum_{h=1}^{G}\pi_h\,\tilde{p}_{h}(z_i;r_i)}, & g\neq 0.
\end{cases}
\end{equation}
which we refer to as our \emph{label-free geometric evidence}. This evidence will serve as the primary indicator for noisy-label identification and robust training in  Sec. \ref{geometric_cons}. During Stage II, the vMF model is updated using all local samples via soft responsibilities, while the samples identified as clean are used to update the class-to-geometry mapping, as detailed in Appendix Sec.~3.

\subsubsection{Annotated Label-dependent Geometry Evidence.}

The preceding stage maps samples onto a spherical representation manifold, where the refined responsibilities $\tilde{\gamma}_{i,g}$, provide a geometry-driven notion of semantic evidence over components $g \in \{0,\dots,G\}$. To incorporate the observed (potentially noisy) labels in a compatible way, we construct a complementary label-based evidence representation in the \emph{same} geometric component space. Concretely, we treat the label-free semantic components as a stable reference and, for each class \(c\), summarize how its labeled samples are distributed across the semantic components.

Formally, we estimate a class-to-geometry distribution via Dirichlet smoothing (small constant $\eta>0$):
\begin{equation}
    \beta_{c,g}\;=\;
    \frac{\sum_{i:\,\tilde{y}_i=c}\!\tilde{\gamma}_{i,g} + \eta}
         {\sum_{g'=1}^{G}\Big(\sum_{i:\,\tilde{y}_i=c}\!\tilde{\gamma}_{i,g'} + \eta\Big)},
\end{equation}
where $\tilde{y}_i$ denotes the observed label of sample $x_i$, and we explicitly exclude the uniform background component $g=0$ from the sums to avoid contamination by geometry-level outliers. Given the class $c$, the vector $\text{B}_c=\{\beta_{c,1},\dots,\beta_{c, G}\}$ describes how that class is expressed across the semantic geometry.

In this way, $ \beta_{c,g}$ embeds the label $\tilde{y}_i=c$ into the same semantic component space as the geometric responsibilities $\tilde{\gamma}_{i,g}$, aligning class space with the representation geometry and enabling a direct comparison between geometry-based and label-based evidence in subsequent stages.

\begin{table*}[t]
\centering
\caption{Experimental results on CIFAR10 with two label noise patterns. The \textbf{bold} denotes the best result while the \underline{underlined} denotes the second-best result.}
\label{tab:cifar10}
\setlength{\tabcolsep}{5pt}
\begin{adjustbox}{width=\linewidth}
\begin{tabular}{lcccccccccccc}
\toprule \toprule
\footnotesize\textbf{Noise Type} & \multicolumn{6}{c}{\textbf{Symmetric Label Noise}} & \multicolumn{6}{c}{\textbf{Pairflip Label Noise}} \\ 
\midrule
\footnotesize\textbf{Noise Pattern} & \multicolumn{3}{c}{\textbf{Localized}} & \multicolumn{3}{c}{\textbf{Globalized}} & \multicolumn{3}{c}{\textbf{Localized}} & \multicolumn{3}{c}{\textbf{Globalized}} \\ 
\midrule
\textbf{Metric} & \textbf{Accuracy} & Precision & F-score & \textbf{Accuracy} & Precision & F-score & \textbf{Accuracy} & Precision & F-score & \textbf{Accuracy} & Precision & F-score \\ 
\midrule

FedAvg \cite{fedavg}  & 34.20 & 46.17 & 30.96 & 42.58 & 42.64 & 41.57 & 44.44 & 50.46 & 41.43 & 43.89 & 45.50 & 42.39 \\
FedProx \cite{fedprox}  & 38.88 & 62.70 & 34.34 & 46.18 & 47.16 & 44.48 & 49.65 & 57.35 & 46.42 & 46.62 & 49.22 & 44.83 \\
MOON \cite{moon}     & 44.88 & 45.73 & 43.20 & 44.88 & 45.73 & 43.20 & \underline{51.68} & \underline{58.97} & \underline{48.80} & 47.54 & 48.76 & 46.22 \\
CCVR \cite{luo2021no}      & 42.63 & 42.44 & 41.53 & 42.63 & 42.44 & 41.53 & 45.16 & 53.53 & 42.30 & 43.69 & 45.20 & 41.95\\
FedDisco \cite{disco} & 43.07 & 43.19 & 41.85 & 43.07 & 43.19 & 41.85 & 44.71 & 49.47 & 41.58 & 43.59 & 45.30 & 42.22 \\
FedSAM \cite{fedsam}  & 41.05 & 60.13 & 37.35 & 46.63 & 46.75 & 45.53 & 50.69 & 58.41 & 43.45 & \underline{59.03} & \underline{64.00} & \underline{54.80} \\
FedExp \cite{fedexp}   & 42.06 & 42.04 & 41.05 & 42.06 & 42.04 & 41.05 & 45.22 & 52.48 & 42.31 & 42.33 & 44.15 & 40.54 \\
FedInit \cite{sun2023understanding} & 36.22 & 48.86 & 33.54 & 42.42 & 42.78 & 41.44 & 46.33 & 53.86 & 43.51 & 42.00 & 43.80 & 40.33 \\
\midrule
SymmetricCE \cite{symmetricCE} & 55.01 & \underline{63.90} & \underline{49.78} & \underline{59.23} & \underline{64.55} & \underline{54.38} & 45.50 & 51.95 & 41.04 & 48.25 & 48.31 & 44.74 \\
FedLSR \cite{FedLSR}     & \underline{55.32} & 57.45 & 48.96 & 55.32 & 57.45 & 48.96 & 35.12 & 25.86 & 25.94 & 35.21 & 24.93 & 25.66 \\
FedNoRo \cite{FedNoRo}   & 43.44 & 43.46 & 42.42 & 40.13 & 40.41 & 39.33 & 46.67 & 48.41 & 46.34 & 43.73 & 43.67 & 42.90 \\
FedELC \cite{fedelc}    & 40.83 & 42.33 & 39.66 & 47.64 & 56.57 & 45.76 & 45.94 & 53.19 & 44.18 & 54.31 & 60.11 & 52.61  \\
FedCorr \cite{fedcorr}   & 39.18 & 44.51 & 42.57 &  35.83 & 48.79 & 41.05 & 40.39 & 39.01 & 37.68 & 36.82 & 34.24 & 41.34 \\
FedClean \cite{jiangfedclean}  & 45.82 & 42.33 & 43.55 & 39.87 & 38.66 & 41.25 & 47.82 & 46.53 & 43.05 & 41.66 & 42.61& 42.88\\

\midrule
\textbf{\method} 
 & \textbf{59.99} & \textbf{70.32} & \textbf{55.31}
 & \textbf{63.29} & \textbf{72.94} & \textbf{59.14}
 & \textbf{61.52} & \textbf{73.01} & \textbf{57.09}
 & \textbf{64.88} & \textbf{64.88} & \textbf{60.98} \\

\bottomrule
\end{tabular}
\end{adjustbox}
\end{table*}

\subsubsection{Geometric Consistency for Noisy Detection}
\label{geometric_cons}
We have obtained two sources of evidence: the geometry-driven evidence from the feature representations, and the label-driven evidence from the observed (potentially noisy) labels. Here, we assess the consistency between these two types of evidence within the same geometric space. Specifically, we examine how well the label information aligns with the geometric representation of each sample.

Given the class-to-geometry distribution, the geometry-label consistency of sample $i$ with its observed label $\tilde y_i$ is computed as an inner product between the class-specific geometry template and the non-background responsibilities:
\begin{equation}
\tilde q_i(\tilde y_i)=\langle \bar{\Gamma}_i,\; \mathbf{B}_{\tilde y_i}\rangle,
\end{equation}
where $\bar{\Gamma}_i=\{\tilde{\gamma}_{i,1},\ldots,\tilde{\gamma}_{i,G}\}$ denotes the distribution of responsibilities for $z_i$. A larger $\tilde q_i $ value indicates that the clusters which \emph{geometrically} explain $z_i$ are also \emph{semantically} aligned with its observed class $\tilde{y}_i$, whereas smaller values flag potential mislabels. Then, we compute a geometry-based cleanliness score as
\begin{equation}
P_i^{clean}
=
\tilde{q}_i(\tilde{y}_i)
=
\sum_{g=1}^{G}\beta_{\tilde{y}_i,g}\,\tilde{\gamma}_{i,g}.
\label{eq:clean_score}
\end{equation}
Larger background responsibility implicitly reduces this score by shrinking the semantic mass assigned to the vMF components. Following prior works~\cite{fedcorr,fedrom}, we fit a two-component Gaussian mixture model (GMM) on $1-P_i^{clean}$ to divide the local samples into a likely-clean subset $\mathcal{D}_{k}^c$ and a likely-noisy subset $\mathcal{D}_{k}^n$.

This geometry-based noisy detection is performed locally within each client during Stage II of every communication round. The corresponding vMF state and the class-to-geometry matrix $B$ are retained on-device across rounds, rather than being transmitted to the server for global aggregation; this preserves client-specific noise characteristics and supports personalized clean/noisy discrimination under heterogeneous FL.

\subsection{Overview of Robust Optimization}
\paragraph{Geometry-model Optimization.}

We initialize the vMF model and the class-to-geometry matrix randomly at the beginning of Stage II. During local client training, we update the vMF model using all local samples to capture the evolving feature geometry, while the class-to-geometry matrix is updated only with the samples identified as clean. Implementation details for these updates are provided in Appendix Sec.~3.

\paragraph{Global Model Optimization.}

The total training objective combines the batch-level robust supervised loss for supervised learning with the forward corrected loss for noise absorption:
\begin{equation}
    \mathcal{L}
=\lambda_s\mathcal{L}_{\mathrm{SCE}}
\ +\ \lambda_{\mathrm{n}} \mathcal{L}_{\mathrm{n}}.
\end{equation}
Here, $\lambda_s$ and $\lambda_{\mathrm{n}}$ are the trade-off hyperparameters. $\mathcal{L}_{\mathrm{SCE}}$ is the noise-resilient loss symmetric cross-entropy (SCE)~\cite{symmetricCE} over all local samples. To explicitly model noisy samples, we introduce a noise absorption matrix $\mathbf{T}\in\mathbb{R}^{C\times C}$, where $T_{c,c'}\approx P(\tilde{y}=c'\!\mid y^{\star}=c)$. Following prior work \cite{10497879}, $\mathbf{T}$ is implemented as an additional linear layer appended after the classifier head. It does not require the model to predict the unknown true label directly. Instead, the classifier outputs the label distribution $p_i$, and the noise absorption layer maps it to the observed noisy-label space via the forward correction $p_i \mathbf{T}$. Concretely, we maximize the likelihood of the observed labels under the forward-corrected distribution:
\begin{equation}
\mathcal{L}_n
=
-\frac{1}{\sum_{i=1}^{B} m_i^{\mathrm{noisy}} + \epsilon}
\sum_{i=1}^{B}
m_i^{\mathrm{noisy}}
\log \bigl[({p}_i \mathbf{T})_{\tilde y_i} + \epsilon \bigr],
\end{equation}
where ${p}_i$ is the classifier output distribution for sample $i$, $\mathbf{T}\in\mathbb{R}^{C\times C}$ is the client-specific noise absorption matrix, $\tilde y_i$ is the observed noisy label, $m_i^{\mathrm{noisy}}\in\{0,1\}$ indicates whether sample $i$ is assigned to the noisy subset by the GMM-based partition, and $\epsilon > 0$ is a small constant. 
\section{Experimental Results}
Our evaluation reports the main results and targeted ablations. A detailed cost analysis and additional experiments are provided  in the Appendix Sec.4 and 5. 
\subsection{Experimental Details} 
\begin{table*}[t]
\centering
\caption{Experimental results on SVHN and CIFAR100 with globalized noise setting. The \textbf{bold} denotes the best result while the \underline{underlined} denotes the second-best result. Localized results are additionally reported in Appendix Sec.~5.1. }
\label{tab:cifar100_SVHN}
\setlength{\tabcolsep}{5pt}
\begin{adjustbox}{max width=\linewidth}
\begin{tabular}{lcccccccccccc}
\toprule \toprule
\textbf{Dataset} & \multicolumn{6}{c}{\textbf{SVHN}} & \multicolumn{6}{c}{\textbf{CIFAR100}} \\
\midrule
\textbf{Noise Type} & \multicolumn{3}{c}{Symmetric Label Noise} & \multicolumn{3}{c}{Pairflip Label Noise}
                 & \multicolumn{3}{c}{Symmetric Label Noise} & \multicolumn{3}{c}{Pairflip Label Noise} \\
\midrule
Metric & Accuracy & Precision & F-score & Accuracy & Precision & F-score & Accuracy & Precision & F-score & Accuracy & Precision & F-score \\
\midrule

FedAvg \cite{fedavg} & 43.52 & 42.79 & 42.29 & 45.73 & 41.04 & 38.08 & 39.27 & 40.81 & 39.53 & 41.49 & 38.79 & 40.53 \\
FedProx \cite{fedprox} & 46.89 & 53.47 & 46.41 & 51.86 & 43.44 & 48.73 & 34.98 & 40.16 & 35.03 & 40.95 & 34.32 & 39.83 \\
MOON \cite{moon}   & 45.15 & 44.79 & 43.51 & 46.10 & 42.55 & 41.11 & 36.57 & 39.78 & 36.69 & 40.33 & 36.14 & 39.45 \\
CCVR \cite{luo2021no} & 45.84 & 41.20 & 45.64 & 44.10 & 42.85 & 35.95 & 39.03 & 41.29 & 38.96 & 42.37 & 38.51 & 40.81 \\
FedDisco \cite{disco} & 43.11 & 41.89 & 43.37 & 43.94 & 40.42 & 36.91 & 38.11 & 40.87 & 38.22 & 41.45 & 37.54 & 40.57 \\
FedSAM \cite{fedsam} & 31.09 & \underline{71.31} & 25.59 & \underline{75.25} & 15.41 & \underline{65.69} & 42.59 & \underline{44.61} & 42.68 & \underline{45.64} & 41.62 & \underline{44.29} \\
FedExp \cite{fedexp} & 46.23 & 41.34 & 45.20 & 43.78 & 43.55 & 66.69 & 38.93 & 41.51 & 39.04 & 41.95 & 38.51 & 41.20 \\
FedInit \cite{sun2023understanding}  & 44.75 & 40.42 & 43.86 & 43.70 & 42.00 & 37.05 & 38.29 & 40.46 & 38.21 & 41.08 & 37.68 & 40.19 \\
\midrule
SymmetricCE \cite{symmetricCE} & \underline{57.82} & 50.69 & \underline{55.28} & 52.28 & \underline{50.56} & 45.71 & \textbf{56.36} & 39.25 & \textbf{58.86} & 40.52 & \textbf{55.80} & 38.65 \\

FedLSR \cite{FedLSR} & 55.21 & 11.07 & 53.07 & 12.16 & 9.88 & 10.54 & 16.14 & 18.72 & 6.57 & 8.17 & 8.26 & 10.23 \\
FedNoRo \cite{FedNoRo} & 37.07 & 39.59 & 34.70 & 39.29 & 34.19 & 37.27 & 34.90 & 39.29 & 34.95 & 40.07 & 34.43 & 39.19 \\
FedELC \cite{fedelc} & 54.24 & 42.72 & 53.79 & 48.69 & \textbf{50.70} & 43.06 & 41.75 & 39.93 & 41.89 & 37.10 & 40.83 & 35.97 \\
FedCorr \cite{fedcorr}  & 49.53 & 42.33 & 48.60 & 43.98 & 45.22 &  39.84  & 36.75 &  39.55  &  40.05 & 40.25 &42.67 & 38.48\\
FedClean \cite{jiangfedclean} & 54.28 & 40.77 & 52.69 & 44.66 & 47.77 & 42.16 & 39.57& 38.88 & 40.22 & 36.71 & 39.83 & 36.79 \\

\midrule

\textbf{\method} 
 & \textbf{62.70} & \textbf{74.55} & \textbf{58.83}
 & \textbf{80.62} & 42.88 & \textbf{68.83}
 & \underline{51.95} & \textbf{56.49} & \underline{54.31}
 & \textbf{60.80} & \underline{51.25} & \textbf{55.46} \\

\bottomrule
\end{tabular}
\end{adjustbox}
\end{table*}

\textbf{Datasets, Models, and Metrics:}  
We evaluate our method on three representative datasets: CIFAR-10 \cite{cifar10}, SVHN \cite{svhn}, and CIFAR-100 \cite{cifar10}. Among them, CIFAR-100 \cite{cifar10} contains richer category diversity with 100 fine-grained classes.
We use ResNet-18 on CIFAR-10, SVHN, and ResNet-34 on CIFAR-100, and keep the backbone setting consistent across all methods. 
We report multiple evaluation metrics, including accuracy, precision, and F-score, following prior works \cite{fedelc, co-teaching, FedLSR}.  Additionally, we include the Clean–Noise Recognition Accuracy (CRA), which measures how accurately the model distinguishes clean samples from noisy ones. Specifically, CRA is computed as the proportion of samples whose predicted clean/noisy status matches the ground truth, reflecting the model’s ability to correctly identify both clean and noisy data instances.

\textbf{Baselines:}
We compare the proposed \method~with methods: \ding{182} Representative FL methods: FedAvg \cite{fedavg}, FedProx \cite{fedprox}, MOON \cite{moon}, CCVR~\cite{luo2021no}, FedDisco \cite{disco}, FedSAM \cite{fedsam}, FedExp \cite{fedexp}, FedInit~\cite{sun2023understanding}. \ding{183} Representative noise-robust methods: Symmetric CE \cite{symmetricCE}, FedCorr \cite{fedcorr}, FedClean \cite{jiangfedclean}, FedLSR \cite{FedLSR}, FedELC \cite{fedelc} and FedNoRo \cite{FedNoRo}. We further compare \method ~with four representation-based methods FedProto \cite{fedproto}, FedCNI \cite{fedcni}, GGEUR \cite{ggeur}, FedROM \cite{fedrom} in Appendix Sec.5.4. 

\subsection{Implementation Details}
\label{implement_details}
\textbf{Federated Learning Setup:} Following previous works~\cite{tang2022virtual,luo2021no}, we conduct experiments under a strongly non-IID setting, where the client data is partitioned according to a Dirichlet distribution with concentration parameter $\alpha = 0.1$. Unless otherwise stated, we adopt $K=10$ clients for all datasets. More details about the hyperparameters are provided in Appendix Sec.1.  

\textbf{Federated Label Noise Setup:}
Following existing methods~\cite{FedNoRo}, we keep the noise rate fixed across all clients and then corrupt the local labels of each client using a transition matrix with noise level $\epsilon=0.4$. In the \emph{globalized} corruption setting, all clients share a common transition matrix, i.e., $p(\tilde{y}=j \mid y=i)$. In the \emph{localized} setting, each client maintains its localized transition matrix that operates within its local label support. For both settings, we consider two standard noise types: (i) \emph{symmetric} and (ii) \emph{pairflip}.

\begin{figure*}[t]
    \centering
    \includegraphics[width=\linewidth]{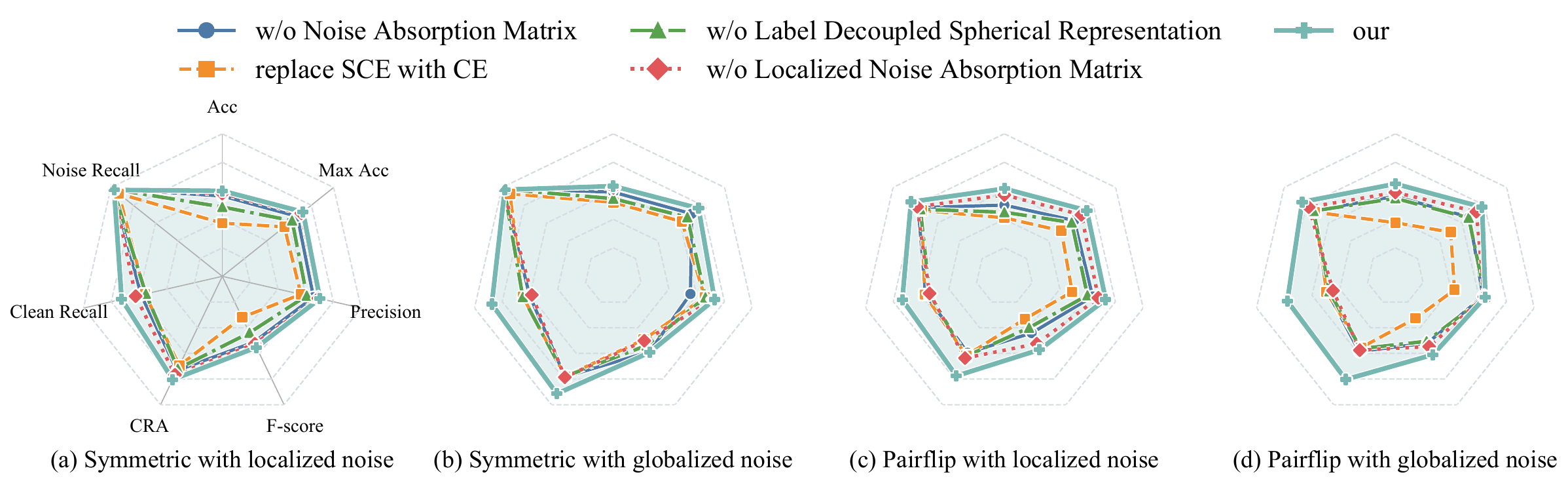}
    \vspace{-0.6cm}
    \caption{
        Comparison of the performance among different schemes under four heterogeneous noise scenarios on CIFAR10.
        Each radar chart corresponds to a specific type of label noise:
        (a) symmetric with localized noise, 
        (b) symmetric with globalized noise, 
        (c) pairflip with localized noise, and 
        (d) pairflip with globalized noise.
    }
    \label{fig:teaser}
    \vspace{-0.6cm}
\end{figure*}


\subsection{Main Results}
Tabs.~\ref{tab:cifar10} and~\ref{tab:cifar100_SVHN} summarize the overall performance of different baselines under both symmetric and pairflip label-noise settings across CIFAR10, SVHN, and CIFAR100. Classical FL algorithms such as FedAvg, FedProx, and MOON degrade significantly under label corruption. Representative noise-robust methods remain competitive under certain settings, but their performance is typically less stable across datasets, noise patterns, and metrics.

In contrast, \method ~consistently achieves the best or second-best results in most reported settings. For CIFAR10 (Tab~\ref{tab:cifar10}), \method ~surpasses all baselines under both localized and globalized settings across symmetric and pairflip noise, especially in accuracy and F-score. For instance, \method ~improves globalized symmetric accuracy by a large margin over the second-best method. For SVHN (Tab.~\ref{tab:cifar100_SVHN}), \method ~is especially strong under globalized pairflip noise, reaching 80.62 accuracy and 68.83 F-score. Some competing methods achieve high accuracy in isolated cases, such as FedSAM, but suffer severe metric imbalance, whereas FedRG remains stronger overall.  For CIFAR100, we observe that SymmetricCE is competitive on a subset of metrics in several settings, especially when the noise pattern is more regular and globally shared. However, this advantage is less consistent across the localized pairflip settings (as shown in Appendix Sec. 5.1).

Overall, these results validate the effectiveness of \method ~in handling heterogeneous and severe label noise in federated environments. Its consistent superiority across datasets and noise types suggests that the geometry-based noise identification benefits robust training.

\subsection{Ablation Study}
\label{ablation_study}
As illustrated in Fig.~\ref{fig:teaser}, we compare \method~against four ablated variants: \ding{182} w/o Noise Absorption Matrix, which removes the forward noise absorption matrix; \ding{183} w/o Label Decoupled Spherical Representation, which eliminates the unsupervised SimCLR in stage I; \ding{184} replaces the SCE loss with a standard CE loss; and \ding{185} w/o Localized Noise Absorption Matrix, which enables the aggregation of the personalized noise absorption matrix across clients.

\begin{figure}[t!]
    \centering

    \begin{subfigure}{0.48\linewidth}
        \centering
        \includegraphics[width=\linewidth]{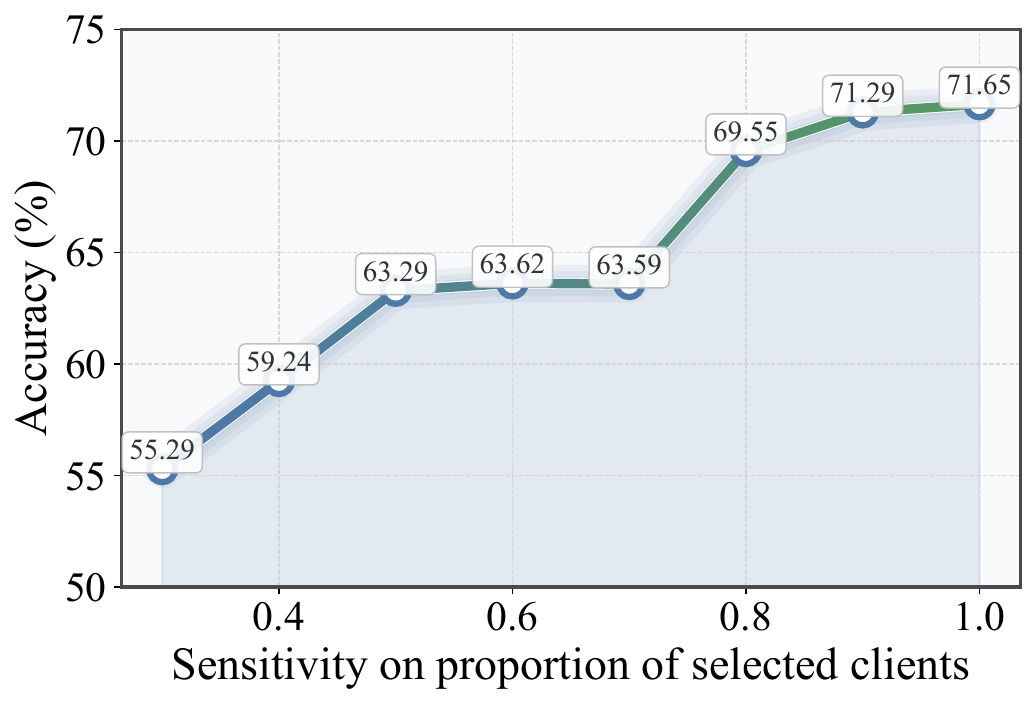}
        \label{ablation_on_num_clients}
    \end{subfigure}
    \hfill
    \begin{subfigure}{0.48\linewidth}
        \centering
        \includegraphics[width=\linewidth]{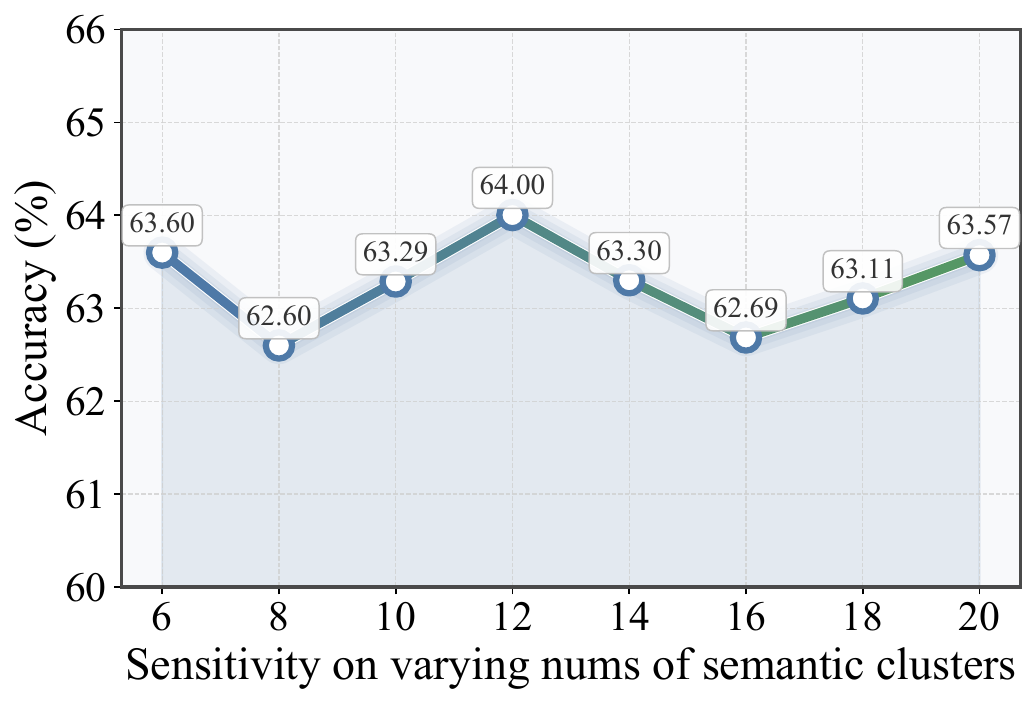}
        \label{ablation_on_num_clusters}
    \end{subfigure}
    \vspace{-0.4cm}
    \caption{
        Comparison of model performance under different numbers of selected clients and semantic clusters.
    }
    \label{fig:two_ablation}
    \vspace{-0.5cm}
\end{figure}

Removing the forward noise-absorption matrix causes a clear drop across almost all metrics, especially in Clean Recall and CRA. This confirms the importance of using the transition matrix to suppress the influence of noisy labels. Second, the removal of the label decoupled spherical representation stage also leads to noticeable degradation. In the early rounds, clients receive highly corrupted supervision. Thus, the unsupervised stage provides noise-free representation learning that significantly benefits downstream robustness. Third, replacing SCE with standard CE indicates that SCE is essential for mitigating gradient bias caused by incorrect labels and preventing overfitting to noise. Finally, aggregating the local noise absorption matrix also lowers performance, indicating that preserving client-specific noise absorption is beneficial. 
Across all four scenarios,  \method ~consistently achieves the best performance. The consistent gaps between \method~and its variants demonstrate that each component contributes meaningfully to the overall robustness to both noise patterns.

We further perform ablation studies about the client sampling rate at each round and the number of clusters $G$ on CIFAR10 with globalized symmetric noise, the results are listed in Fig.~\ref{fig:two_ablation}. The performance of \method~continues to improve with the increase in the number of participating clients in each round. The selection of the number of semantic clusters is also a key parameter for \method. In our other experiments, we set the number of clusters to 10. In addition, we also test the impact of different numbers of semantic clusters ranging from 6 to 20. It can be seen that the performance remains balanced overall and is robust to the number of semantic clusters in the reported cases.

\section{Conclusion}
We introduce \method~ to address the challenges of noisy labels under data heterogeneity by the representation geometry principle over traditional loss-based methods. \method~robustly identifies noisy samples through geometric consistency and employs a personalized noise absorption matrix for optimization. Extensive experiments demonstrate that \method~outperforms the state-of-the-art baselines.

\textbf{Limitations and Future Directions:} FedRG currently focuses on image classification, leaving graph and natural language domains unexplored. As the first work on OOD generalization on graphs~\cite{li2022ood} demonstrates, structural distribution shift poses unique challenges; future work could extend our geometric framework to such settings. Additionally, leveraging the first benchmark and library for curriculum learning~\cite{zhou2024curbench, zhou2022curml}, incorporating sample difficulty into our clean/noisy partition could enable finer-grained robust training in heterogeneous FL scenarios.

\section*{Acknowledgements} Tian Wen, Xuefeng Jiang and Yuwei Wang were supported by the National Key Research and Development Program of China (Grants No. 2023YFB2703700). Zhiqin Yang, Yonggang Zhang, and Bo Han were supported by NSFC General Program No. 62376235 and RGC GRF No. 12200725. Yonggang Zhang was also funded by Inno HK Generative AI R\&D Center. Hao Peng was supported by the Local Science and Technology Development Fund of Hebei Province Guided by the Central Government of China, through grant 254Z9902G.
\bibliographystyle{unsrt}
\bibliography{main}
\end{document}